\documentclass[conference]{IEEEtran}
\IEEEoverridecommandlockouts
\pdfoutput=1

\usepackage{cite}
\usepackage{amsmath,amssymb,amsfonts}
\usepackage{algorithmic}
\usepackage{graphicx}
\usepackage{textcomp}
\usepackage{url}
\usepackage[english]{babel}
\usepackage[utf8]{inputenc}
\usepackage[T1]{fontenc}
\usepackage{amsthm}
\usepackage{float}
\usepackage{tabularx}

\usepackage{orcidlink}

\def\BibTeX{{\rm B\kern-.05em{\sc i\kern-.025em b}\kern-.08em
    T\kern-.1667em\lower.7ex\hbox{E}\kern-.125emX}}

\usepackage{color}
\definecolor{itwmgreen}{RGB}{23,156,125}
\definecolor{itwmblue}{RGB}{0,110,146}
\definecolor{itwmlightblue}{RGB}{212,230,244}
\definecolor{itwmviolett}{RGB}{57,55,139}
\definecolor{itwmorange}{RGB}{235,106,10}

\usepackage{listings}
\begin{document}

\title{Python Workflows on HPC Systems}
\author{
  \IEEEauthorblockN{Dominik Strassel$^{1,\orcidlink{0000-0003-3527-9745}}$, Philipp Reusch$^{1,\orcidlink{0000-0002-7532-3782}}$ and Janis Keuper$^{2,1,\orcidlink{0000-0002-1327-1243}}$ }
  \vspace{0.5cm}
  \IEEEauthorblockA{\it{$^1$CC-HPC, Fraunhofer ITWM, Kaiserslautern, Germany}}
  \IEEEauthorblockA{\it{$^2$Institute for Machine Learning and Analytics, Offenburg University,Germany}}
  \thanks{Electronic address: dominik.strassel@itwm.fraunhofer.de}
  \thanks{Submitted and accepted at the \href{https://pyhpc.io/}{PyHPC Workshop} at \href{https://sc20.supercomputing.org/}{SuperComputing 2020} and will be published at IEEE TCHPC Proceedings. The copyright is finally transferred after publication.}
  \thanks{\copyright 202X IEEE. Personal use of this material is permitted. Permission from IEEE must be obtained for all other uses, in any current or future media, including reprinting/republishing this material for advertising or promotional purposes, creating new collective works, for resale or redistribution to servers or lists, or reuse of any copyrighted component of this work in other works.}
}

\maketitle

\begin{abstract}
The recent successes and wide spread application of compute intensive machine learning and data analytics methods have been boosting the usage of the \texttt{Python} programming language on HPC systems. While \texttt{Python} provides many advantages for the users, it has not been designed with a focus on multi-user environments or parallel programming - making it quite challenging to maintain stable and secure \texttt{Python} workflows on a HPC system. In this paper, we analyze the key problems induced by the usage of \texttt{Python} on HPC clusters and sketch appropriate workarounds for efficiently maintaining multi-user \texttt{Python} software environments, securing and restricting resources of \texttt{Python} jobs and containing \texttt{Python} processes, while focusing on Deep Learning applications running on GPU clusters.
\end{abstract}
\vspace{0.25cm}

\section{Introduction}\label{sec_intro}
In recent years, \texttt{Python} has become the most popular programming language~\cite{stackoverflow} and there are many reasons why \texttt{Python} has evolved from a simple scripting language into the dominating implementation environment: it is easy to learn, platform independent, supports the full range of programming paradigms (from functional to full object oriented), it allows easy integration of external software modules (allowing performance critical parts to be written in other languages) and for most, it comes with an exhaustive ecosystem of available open source software libraries. But \texttt{Python} also has a ``dark'' side. Originally, it was not designed with a focus on multi-user, shared resource systems or to be used for massive parallel processing. While users and developers have come up with many ways to work around these initial design choices over the last years, some core properties of \texttt{Python} (and some of the workarounds) are still prone to cause massive problems in the maintenance and operation of HPC systems. In result, we, as many others in the HPC community, have been experiencing a wide range of small and severe issues related to \texttt{Python} workflows on various HPC systems. Typical examples are: 
\begin{itemize}
\item \texttt{Python} jobs tend to spin off a vast amount of child processes, some are not terminating after a job ends. A phenomenon that regularly causes following GPU-jobs to crash when these processes are still occupying GPU memory.
\item \texttt{Python} jobs appear to be ``escaping'' the resource control mechanisms of batch systems on a regular basis, allocating more memory and CPU cores than scheduled - hence, affecting other users on the system.
\item The maintenance and management of the diverse and fast evolving \texttt{Python} software environment is quite challenging, especially when the needs of a diverse user group are contradicting.
\end{itemize}

This paper presents the intermediate results of our ongoing investigation of causes and possible solutions to these problems in the context of machine learning applications on GPU-clusters. In section~\ref{sec:resource-management} we address the necessity of restrictive job termination in the context of \texttt{Python} based algorithms on multi-user HPC systems. Followed by section~\ref{sec:controlling-python}, where we discuss in detail how to control and restrict ressources aquired by those \texttt{Python} processes. In section~\ref{sec:python-environment} we show some possibilities to create and maintain \texttt{Python} environments.

\section{Safely Terminating Python Processes}\label{sec:resource-management}
Correlating with the increasing use of \texttt{Python} workflows, we\footnote{And many forum posts suggest that this is a wide spread problem} observed a growing number of incidents, where different batch systems on several HPC-Clusters were not able to sufficiently terminate jobs running python scripts. While these problems were quite hard to track and to isolate, they all shared the common pattern, that a child process triggered by a python script did not terminate with the associated job. Especially on GPU-Clusters, where machine learning applications require a lot GPU memory, these renegade processes cause severe problems to following GPU jobs of other users.

In order to analyze the details of this problem (in section~\ref{sec_1_sol}), we first need to review the intrinsics of child processes and their termination in \textit{Linux} (section~\ref{linux_child}) and the dynamics of child process generation in  \texttt{Python} (section~\ref{python_child}).

\subsection{Terminating Child Processes in Linux}\label{linux_child}
The spawning of child processes is (along with multi-threading) the standard parallelization technique provided by the \textit{Linux} kernel. Hence, one should assume, that the usage of child processes is a safe and deterministic approach. However, there are some pitfalls, which are often caused by the way these processes are handled in the event that their parents terminate.

To illustrate the termination mechanics, we use a \texttt{BASH} script (Algorithm~\ref{lst:child processes}) that generates two child processes that do nothing but \texttt{sleep} for $100$ and $150$ seconds, respectively.

\begin{lstlisting}[language=Bash,caption={Simple BASH script generating two child processes.},captionpos=b,label={lst:child processes}]
#!/bin/bash
set -m
trap 'kill %%' EXIT
sleep 100 &
sleep 150 &
wait
exit 0
\end{lstlisting}

In order to stop this script we have different possibilities, but in the end it always involves sending a signal to the process telling it to end. In the following we focus on only two signals -- \texttt{SIGTERM} and \texttt{SIGKILL}. \texttt{SIGTERM} is the default signal sent when invoking kill, e.g. used by \texttt{top} or \texttt{htop}. It tells the process to shut-down gracefully. This is the signal that should be used if we want that the process to shut down cleanly. The problem is, that technically \texttt{SIGTERM} can be ignored by a process, as it is only informed to shut-down and close all open files, databases and so on. But this termination signal is handled \textit{inside} the process. In most situations it is considered as ``bad practice'' not to terminate after getting a \texttt{SIGTERM} signal. In contrast, \texttt{SIGKILL} cannot be ignored by a process as the termination is handled outside of the process. \texttt{SIGKILL} is useful when an application does not respond any more \textit{or} will not terminate with \texttt{SIGTERM}. In general \texttt{SIGKILL} should be able to stop every process, but there are exceptions. The most prominent example for such an exception are the so-called \textit{``zombie processes''}.

\begin{figure}[t]
  \centering
  \includegraphics[width=\columnwidth]{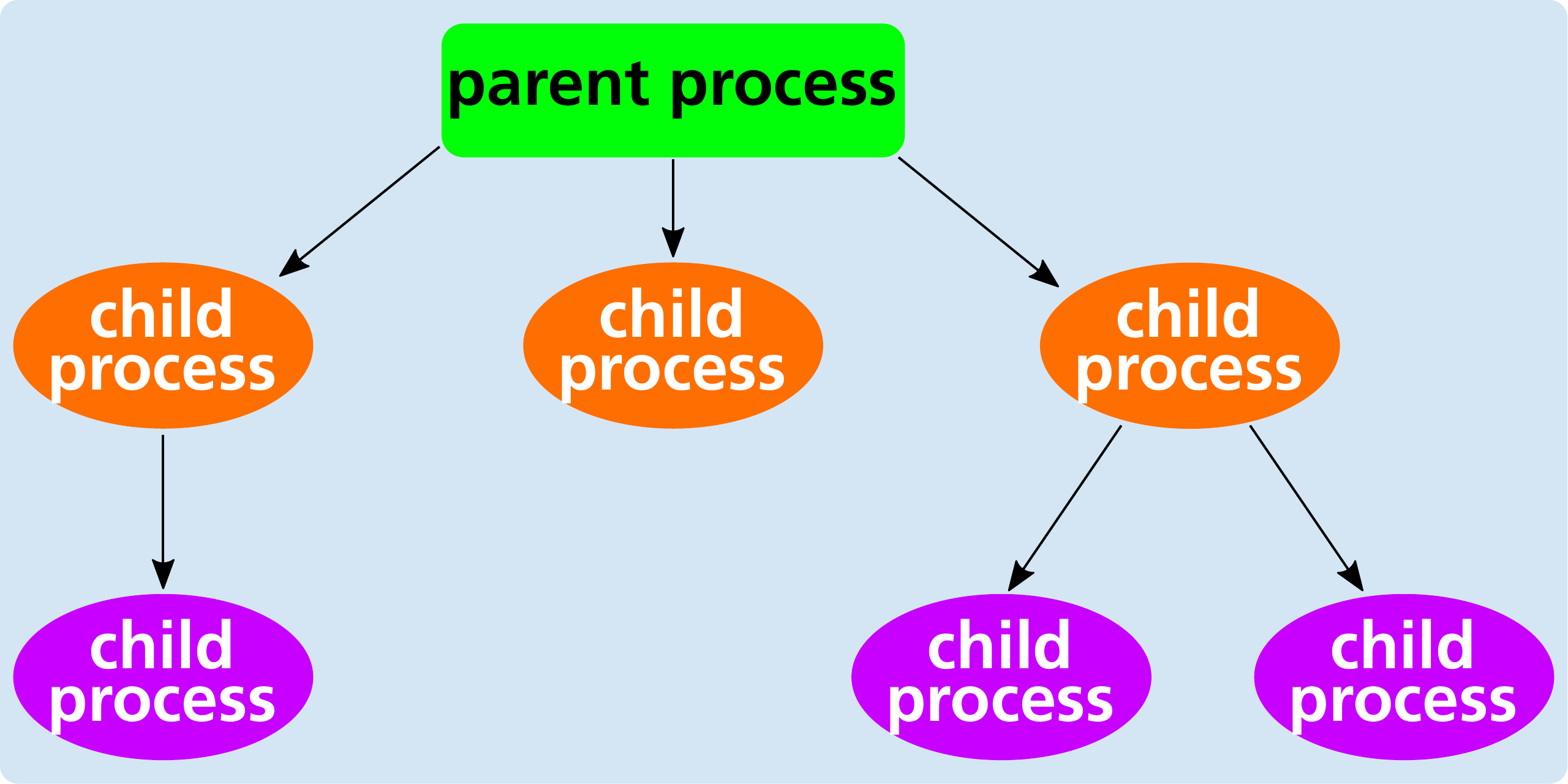}
  \vspace*{-0.25cm}
  \caption{(color online) Example of child process spawning.}
  \label{figure:paralle-compute-python}
\end{figure}

A zombie process is a process that has finished normally and is waiting for its parent to read its exit state. The crucial point is, that the child process has finished its execution, or in other words it is \textit{``dead''}. But the entry in the process table still exists. The most likely situation for a zombie process to appear is when the parent job has died (e.g. segfault or the process got a \texttt{SIGKILL}). Sometimes it is also possible that the parent process is still active (and the parent process has not $\mathrm{PPID} = 1$). This indicates that the parent process is stalled on a task and therefore cannot read the exit state of the child process. On the other side, we have what is called an \textit{``orphaned process''}. This kind of process is very similar to a zombie process. But in contrast to a zombie process, an orphaned process is a process that is still running and whose parent has died. By default, an orphaned process will be adopted by the \textit{init} process which results in a parent-pid of $\mathrm{PPID} = 1$, but is still owned by the user that has started the parent process. Following this definition, every ``daemon process'' is strictly speaking an orphaned process, where the termination of the parent process is done on purpose and not by error or a \texttt{SIGKILL}.

If we stop Algorithm~\ref{lst:child processes} with either \texttt{SIGTERM} \textit{or} \texttt{SIGKILL} the child processes become orphaned processes and are not killed. As these processes continue running, they continue consuming resources. In this simple example this is not problematic, but if we consider more resource intensive scenarios, e.g. a child that allocates GPU memory, we will run into massive problems. Then, the GPU memory -- one of the most limited resources on an HPC cluster -- will  not be freed at the end of a job because the child is not terminated. At this point one would argue that we have a batch system on every HPC cluster and every user program is started as a batch job. Which is true. The problem is that orphaned processes have $\mathrm{PPID} = 1$ and are therefore no longer part of the batch job. This results in the fact that the batch system cannot stop such orphaned processes directly. Hence, Algorithm~\ref{lst:child processes} is a shockingly trivial example of a scenario where child process are actually able to ``escape'' the batch system if no further counter measures are in place.

\subsection{Generation of child processes in Python}\label{python_child}
The problem of terminating child processes in batch system environments is not specific to the execution \texttt{Python} scripts on GPUs - in fact Algorithm~\ref{lst:child processes} is a \texttt{BASH} script running on CPU. However, there are several reasons why these problems correlate with \texttt{Python} machine learning applications on GPU-Clusters: First, GPU memory is (compared to main memory) a precious resource - orphans blocking a few GB of memory will more likely cause problems to following users. Second, users running machine learning applications are often trying to ``work around'' the batch system, by starting interactive environments like \textit{Jupyter}\footnote{Of cause this can't be done without the necessary network configuration, but we see more and more HPC sites allowing their users to use \textit{Jupyter}~\cite{jupyter} within the batch system}~\cite{jupyter}. Within these environments, users then run scripts interactively and tend to terminate them via \texttt{CTRL-C}, producing orphans ``on the fly''. Last, there is the  \texttt{Python} specific, extensive usage of child processes, which makes it simply more likely to end up with orphans compared to languages like \texttt{C}/\texttt{C++}, were most of the intra-node parallelization is implemented via multi-threading.

The usage of multi-threading is the standard in-node parallelization mechanism used on HPC systems and compiled high-level programming language -- like \texttt{Fortran}, \texttt{C} and \texttt{C++}. Due to basic design choices within \texttt{Python}, multi-threading does not work the same way as in these languages. The standard \texttt{Python} implementations\footnote{e.g. \texttt{CPython}} have a global lock on the interpreter~\cite{GIL}. This lock \textit{``...must be held by the current thread before it can safely access Python objects.''}~\cite{GIL2} This remains true, even if we use libraries like \texttt{threading}. Even though we can create multiple threads with this library, the execution is still limited via the global interpreter lock (GIL). Such a behavior is totally different to the way most programmers would understand the concept of ``threading''. In order to get around the limitation of the GIL, \texttt{Python} developers started neglecting parallelization by multi-threading and turned to multi process solutions instead. 
In this approach, jobs spawn several child processes (see Fig.~\ref{figure:paralle-compute-python}), e.g. with \texttt{Pool}~\cite{Pool}, in order to achieve true parallel execution of \texttt{Python} code. Note that this does not violate the GIL as \textit{``...[t]he multiprocessing package offers both local and remote concurrency, effectively side-stepping the Global Interpreter Lock by using subprocesses instead of threads...''}~\cite{Pool}. In combination with the tendency to use many third party libraries, which are spawning their own children, this leads to the notable effect, that many python applications end up with dozens or more child processes.

\subsection{Jailing Processes}\label{sec_1_sol}
In order to overcome the previously described behavior, one has to somehow ``jail'' everything that runs inside the batch jobs. On Unix-like systems one can make use of the \texttt{seccomp}~\cite{seccomp} kernel feature, \texttt{process groups}~\cite{setpgid} (e.g. \texttt{setpgid}, \texttt{setpgrp}) or \texttt{PID namespaces}~\cite{pidnamespace}. This could be done by adjusting the batch system either via integrating the usage of \texttt{seccomp} and \texttt{PID namespaces} in the batch system or the less preferable variant -- forcing users to implement those features in their programs. Besides the ability to suppress escaping a batch job with the help of \texttt{seccomp}, one has the opportunity to limit the folders that are visible or accessible within the batch job. Another quickly implemented possibility is to use already existing tools like \texttt{firejail}~\cite{firejail} or \texttt{newpid}~\cite{newpid}. No matter which solution is going to be the preferred one, it will be unavoidable to implement such a feature in batch systems. Not only because of the fact that we want to avoid orphaned processes, but also with regard to the coming integration of cloud solutions in HPC clusters in the near future.

\section{Controlling the Resources of Python Processes}\label{sec:controlling-python}
The second problem we empirically observed on various HPC clusters, is the lack of resource control. Using batch-systems, one would expect that (\texttt{Python}) processes would not be able to exceed predefined memory and CPU resources. However, in the next sections we discuss a number of direct and indirect effects caused by \texttt{Python} workflows, that can bypass resource limits at the expense of other users on the system.

\subsection{Memory Limits of Batch-Jobs}\label{sec:memory}
The first observation is only indirectly caused by \texttt{Python} and is more likely the result of the high dynamics in the ``\texttt{Python}-GPU-Machine-Learning'' environment. The application driven need to run on the latest GPU drivers (also see section~\ref{sec:python-environment}), forcing HPC system maintainers to use much newer kernel versions than the typically very conservative system management would allow. If not all system components, especially the batch-system, are up to date with newer kernels, we have seen heavy side effects. Most notably the total breakdown of memory limits in batch-systems. The problem is, that the usage of ``\texttt{ulimit -m}'' by older versions of batch-systems, which is supposed to limit the max memory size, has no effects since Linux kernel version 2.4!~\cite{depre-ulimit}. We performed a little test with a simple program to illustrate the effects, see Table~\ref{tab:ulimit-m}.

\begin{table}[t]
  \centering
  {\normalsize
  \begin{tabularx}{\columnwidth}{| p{0.4\columnwidth} | X | c |}
    \hline
    \hline
    \textit{Command} & \textit{Limit} \textit{Size} & 5GB \textit{allocated} \\
    \hline
    \hline
    \texttt{ulimit -m 1} & 1 KB & yes\\
    \texttt{ulimit -m 1048576} & 1 GB & yes\\
    \texttt{ulimit -m 5242880} & 5 GB & yes\\
    \texttt{ulimit -m 20971520} & 20 GB & yes\\
    \texttt{ulimit -m 41943040} & 40 GB & yes\\
    \hline
    \hline
  \end{tabularx}
  }
  \caption{Trying to allocate 5GB of main memory with various memory restrictions using \texttt{ulimit -m}.}
  \label{tab:ulimit-m}
\end{table}

In today's Unix-like systems with activated \textit{control groups}~\cite{cgroups}, this is not a problem as long as the control groups are activated in the batch system as well. But there are some restrictions to this statement. On some Debian derivatives the memory cgroups are excluded and have to be added explicitly to the kernel command line.~\cite{slurm-cgroups} The same yields true for the configuration of batch systems, where the default setting in many cases is still without control groups. It is even the case that some batch systems reimplement the usability of ``\texttt{ulimit -m}'' in such a way, that they check (e.g.) every 30 seconds the used memory of a batch job.~\cite{slurm-conf} This makes it, with some limitations, possible to kill programs that use more memory than the batch job has allocated. But relying on those kinds of reimplementation is dangerous, as these checks rely on a given time interval, in the example 30 seconds. Note that if the process hierarchy in a batch job is too deep this reimplementation does not work either. Combining both makes it possible to kill an entire compute node within less than a few seconds with a simple ``\texttt{tail /dev/zero}''!

Besides control groups, there is another \texttt{ulimit} that can be used to limit the memory usage of program -- ``\texttt{ulimit -d}''. With this one can limit \textit{``...[t]he maximum size of the process's data segment (initialized data, uninitialized data, and heap). This limit affects calls to brk(2) and sbrk(2), which fail with the error ENOMEM upon encountering the soft limit of this resource...''}~\cite{depre-ulimit} To check this we tested this in the same way like before, see Table~\ref{tab:ulimit-d}.

\begin{table}[t]
  \centering
  {\normalsize
    \begin{tabularx}{\columnwidth}{| p{0.4\columnwidth} | X | c |}
      \hline
      \hline
      \textit{Command} & \textit{Limit} \textit{Size} & 5GB \textit{allocated} \\
      \hline
      \hline
      \texttt{ulimit -d 1} & 1 KB & no\\
      \texttt{ulimit -d 1048576} & 1 GB & no\\
      \texttt{ulimit -d 5242880} & 5 GB & no\\
      \texttt{ulimit -d 20971520} & 20 GB & yes\\
      \texttt{ulimit -d 41943040} & 40 GB & yes\\
      \hline
      \hline
    \end{tabularx}
  }
  \caption{Trying to allocate 5GB of main memory with various memory restrictions using \texttt{ulimit -d}.}
  \label{tab:ulimit-d}
\end{table}

And indeed this limit still works. Nevertheless using control groups has much more benefits and has the possibility to act on a different level but ``\texttt{ulimit -d}'' can still be a handy command to know.

\subsection{Pitfalls of Python Memory Management}
Once we are able to fully contain the memory usage of \texttt{Python} jobs, users start to experience yet another, very \texttt{Python} specific problem: the odds of \texttt{Python} memory management. Being an interpreted language that takes all direct memory interaction away from the programmer, the \texttt{Python} memory management makes it hard to estimate the actual memory demands of an algorithm. On top of that, memory consumption often appears to be non deterministic, in the sense that small changes in the code can cause large changes in memory usage, which leaves users with unexpected \textit{out of memory} job terminations.

\texttt{Python} has been designed to abstract various levels of control from the users, including memory management. This enables users with less knowledge in software development to write custom applications, but bears a challenge, if more control is required. While it is theoretically possible to predict the memory usage under certain conditions, it requires deep knowledge of the \texttt{Python} memory management and the used interpreter.

To abstract the memory management from the user, \texttt{Python} creates a private heap for \texttt{Python} objects~\cite{python-memory} and the \texttt{Python} data model defines, that \textit{``[a]ll data [...] is represented by objects or by relations between objects''}~\cite{python-datamodel}. This results in wrapping every value in an object, before storing it in the memory. The overhead of the wrapper can be shown by running the \texttt{Python} code from Algorithm~\ref{lst:pyobjectsize}.

\begin{lstlisting}[language=Bash,mathescape=true,caption={Print the object size of an integer literal},captionpos=b,label={lst:pyobjectsize}]
# Python 3.6.4, manjaro 18.0.4 x64 kernel 4.19.66
>>> import sys
>>> print("integer:", sys.getsizeof(2147483647))
integer: 32
\end{lstlisting}

The ``\texttt{sys.getsizeof()}'' call in Algorithm~\ref{lst:pyobjectsize} \textit{``[r]eturn[s] the size of an object in bytes''}~\cite{python-sys}. The call argument is an integer literal representing the number $2147483647$ and the reported object size is $32$ bytes. In an comparable C application, the number $2147483647$ is the maximum positive value of the ``\texttt{signed integer}'' type, which takes up four bytes. In comparison, the \texttt{Python} object requires eight times the space of the C equivalent. This overhead is introduced by the metadata required for each object. The exact amount of overhead depends on the stored value -- e.g. a small integer like $127$ requires $28$ bytes. In result, the memory complexity depends on the concrete values of the data. Especially in research with huge amounts of heterogeneous data, this can be hard to predict.

Even after determining the maximum size of an object in bytes, the creation of an object is not directly related to memory allocations. \texttt{Python} internally handles the allocation of memory depending on the size of an object. In \texttt{CPython} by default objects larger than $512$ bytes are directly allocated on the application heap, while smaller objects will be stored in pre-allocated arenas \textit{``[...] with a fixed size of $256$ KB''}~\cite{python-memory}. This aggravates the prediction of allocations, as it depends on the size of the object and the current utilization of the arenas.

Similar behavior applies to the deallocation of memory. Non-aggregate objects are freed, when the reference counter for the object is zero. Aggregate data types like dicts, lists or tuples, are watched by the garbage collection (\texttt{gc}) module. The \texttt{gc} stores objects in multiple generations, depending on the age of an object. A generation is a collection of objects with the same age. The age of an object is the number of \texttt{gc} rounds without being freed. Each generation has a counter for the number of allocations minus the number of deallocations since the last \texttt{gc} round. Based on this counter a per-generation threshold can be defined. When the counter exceeds the threshold, a new \texttt{gc} round is initiated. In a \texttt{gc} round, all unused objects of the generation with exceeded threshold are freed and the remaining objects are moved to the next-older generation.~\cite{python-gc} The default values in \texttt{CPython} are three generations with thresholds $g_0=700, g_1=10, g_2=10$ with $g_0$ as youngest generation. Even knowing the parameters, it is really hard to predict the exact time of deallocations, because they highly depend on the object handling of the application.

The overhead introduced by the object pattern also impedes the compute performance. In research with huge amounts of data, it is common to use third-party libraries like \texttt{numpy} or \texttt{scipy} to overcome these performance issues. A major reason for the better performance, is an improved memory management -- e.g. for arrays in \texttt{numpy}. By storing the data in a custom structure, the overhead of the object pattern can be reduced. The custom structure further allows to provide optimized algorithms, which can achieve remarkable performance gains. The downside of the use of third-party libraries is the introduction of unknown memory operations, which requires additional knowledge of the memory management of the used library.

Apart from the mentioned arguments, the final memory management is done by the \texttt{Python} interpreter executing the script in a specific hardware/software environment. Even with the prediction for a specific setup, changing a single part may result in a highly different memory pattern. On HPC-clusters, this leads to a situation in which the cluster requires hard limits, but the users do not have the tools to write \texttt{Python} applications, which guarantee to respect these limits. As a fix it might be considered to provide a test queue with short wall time and high priority. This enables users to trial and error their \texttt{Python} programs to achieve specific utilization goals, before submitting them in the intended queue. Tools like ``\texttt{ulimit}'' or ``\texttt{cgroups}'' can be used to provide the required memory limitation to avoid out-of-memory crashes of compute nodes.

\subsection{Limiting CPU Usage}\label{sec:cpus}
Traditional HPC workflows use mostly multi-threading for in-node parallelization. The threading APIs of compiled high-level programming languages like \texttt{C/C++} or \texttt{Fortran}, the libraries written in these languages and HPC batch systems support this standard approach by honoring given thread limits. These limits are usually set by the batch-script at the start of a job, setting the limits via a few environment variables like \texttt{OMP\_NUM\_THREADS}, which sets the number of threads to be used by \texttt{openMP}, or \texttt{MKL\_NUM\_THREADS}, which sets the same for calls to the \texttt{Intel Math Kernel Library}, just to name a few examples. Unfortunately, \texttt{Python} jobs will often not respond to these mechanism: while pure \texttt{Python} scripts, that are using child process for parallelisation (see section~\ref{python_child}), will simply ignore thread limits and occupy as many cores as they can get by spawning process after process, some \texttt{Python} scripts which are interfacing external libraries, e.g. such as \texttt{NumPy}~\cite{numpy}, might follow the limits to some extent. In the following, we will analyze the threading behavior of three very popular libraries -- \texttt{Tensorflow}~\cite{TensorFlow}, \texttt{PyTorch}~\cite{PyTorch} and \texttt{NumPy} -- which we observed to be commonly used in \texttt{Python} jobs on HPC-systems.

In order to control the appearance of threads we have environment variables like, \texttt{MKL\_THREADING\_LAYER}, \texttt{MKL\_NUM\_THREADS}, \texttt{NUMEXPR\_NUM\_THREADS}, or \texttt{OMP\_NUM\_THREADS}. Normally we can use those variables in a \texttt{BASH} script -- Algorithm~\ref{lst:envtest} -- and we are able to limit the number of threads.

\begin{lstlisting}[language=Bash,mathescape=true,caption={Limiting the number of threads},captionpos=b,label={lst:envtest}]
#!/bin/bash
export MKL_THREADING_LAYER=SEQUENTIAL
export MKL_NUM_THREADS=$\alpha$
export NUMEXPR_NUM_THREADS=$\beta$
export OMP_NUM_THREADS=$\gamma$
COMMAND-TO-EXECUTE
exit 0
\end{lstlisting}

Whereas $\alpha, \beta, \gamma$ are simple numbers larger than zero. We did in total $576$ ($3 \times 192$) simulations with $\alpha, \beta, \gamma = \lbrace 1, 2, 8, 16 \rbrace$ and $20$ CPU cores. The node was used exclusively for the simulations. For the tests we used three different algorithms summarized in Table~\ref{tab:simulation-setup}.

\begin{table}[t]
  \centering
  {\normalsize
  \begin{tabularx}{\columnwidth}{| p{0.16\columnwidth} | p{0.14\columnwidth} | p{0.14\columnwidth} | X |}
    \hline
    \hline
    \textit{Framework} & \textit{Topology} & \textit{Dataset} & \textit{Comment} \\
    \hline
    \hline
    NumPy & -- & -- & {\small mandelbrot calculation with $x \in \left[-2.4, 1.2\right]$, $y \in \left[-1.2, 1.2\right]$, resolution $6400 \times 4800$, $256$ iterations}\\
    Tensorflow & ResNet50 & ImageNet & classification task ($18000$ training steps)\\
    PyTorch & ResNet50 & Voc2012 & segmentation task ($1500$ training steps)\\
    \hline
    \hline
  \end{tabularx}
  }
  \caption{Simulation setup to test Algorithm~\ref{lst:envtest}.}
  \label{tab:simulation-setup}
\end{table}

\begin{figure}[t]
  \centering
  \includegraphics[width=\columnwidth]{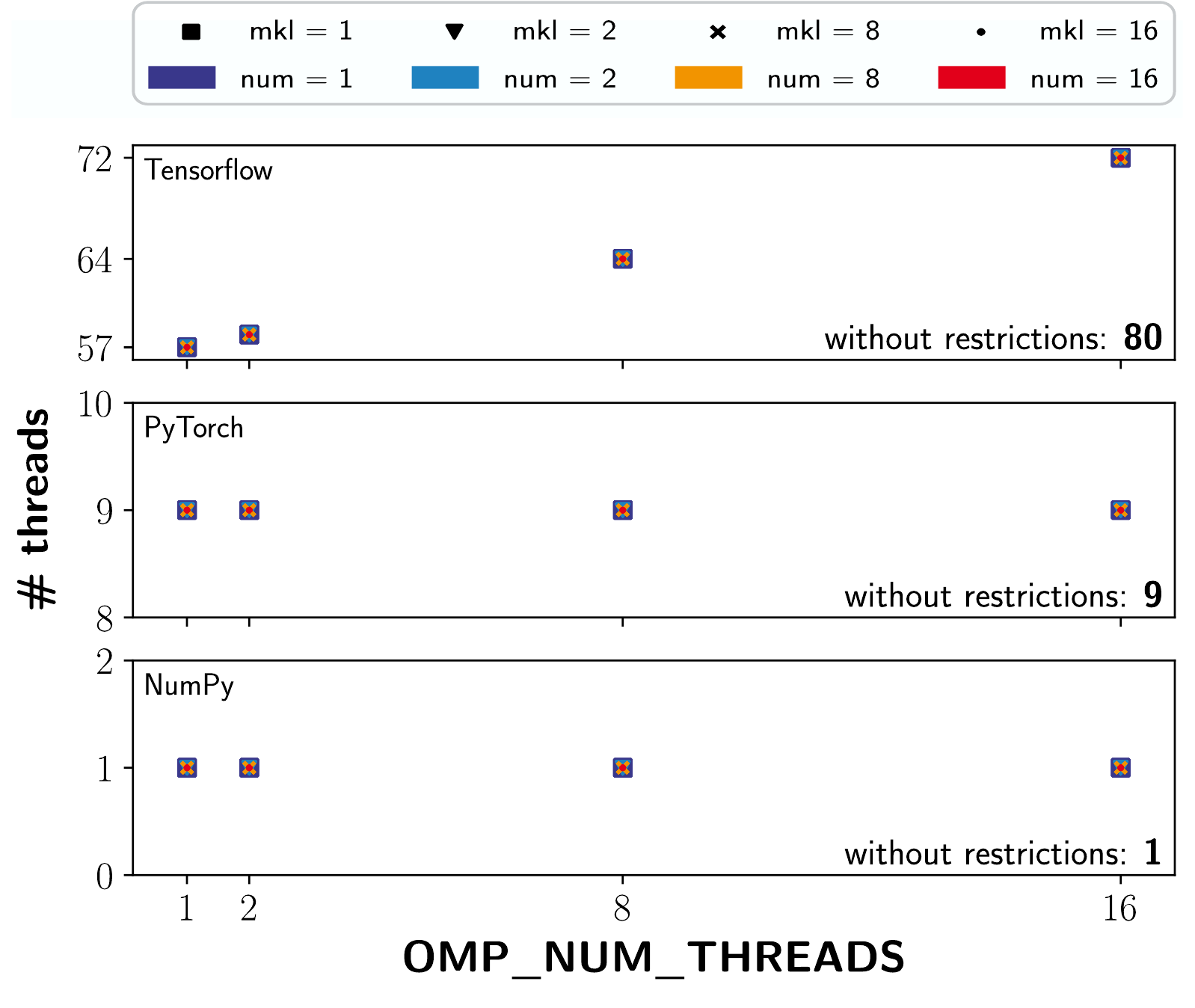}
  \vspace*{-0.25cm}
  \caption{(color online) Number of threads as a function of \texttt{OMP\_NUM\_THREADS} for the TensorFlow (top), PyTorch (middle) and NumPy (bottom) simulations as defined in Table~\ref{tab:simulation-setup}. The remaining environment variables as defined in Algorithm~\ref{lst:envtest} are encoded via colour or symbol type respectively.}
  \label{figure:threads}
\end{figure}

\begin{figure}[t]
  \centering
  \includegraphics[width=\columnwidth]{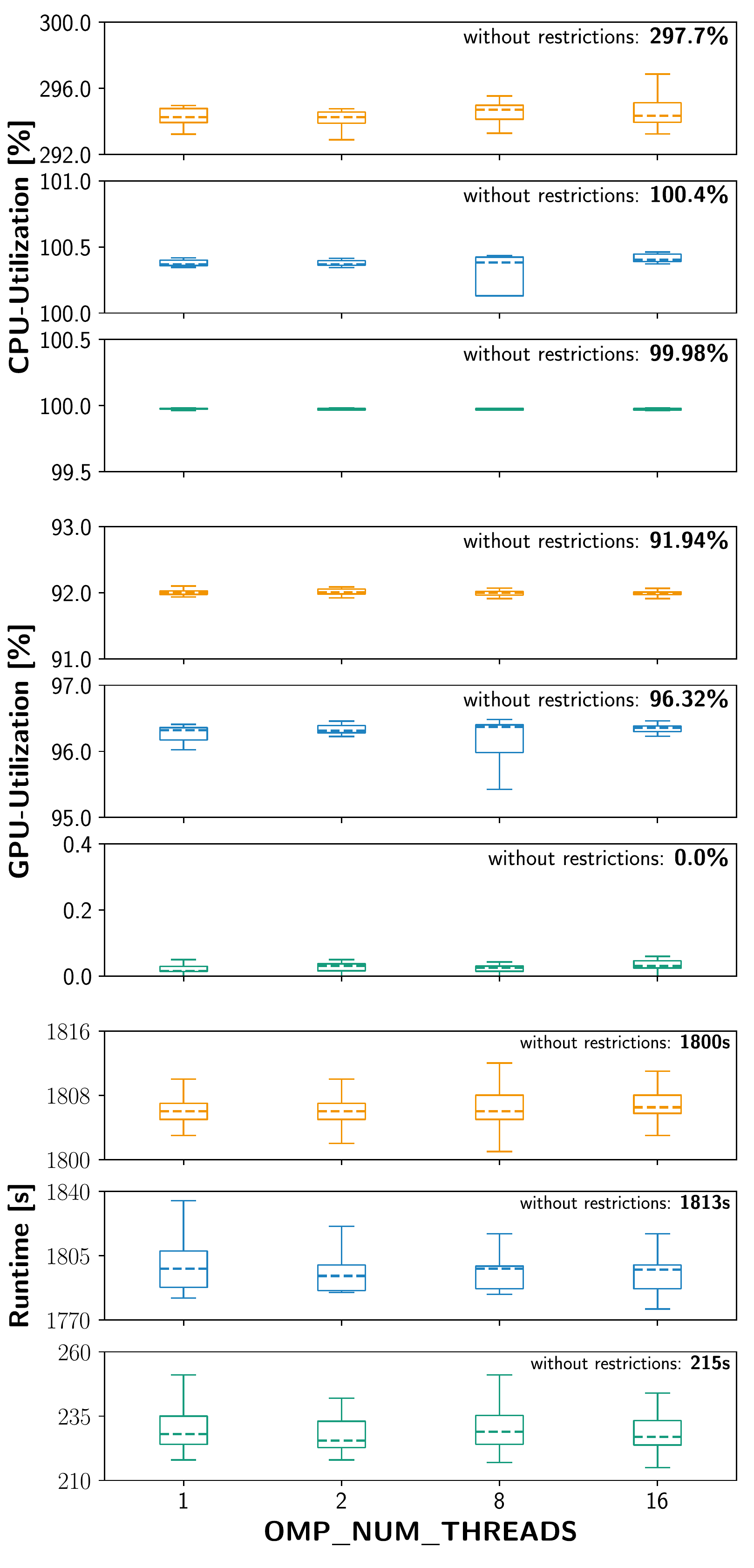}
  \vspace*{-0.25cm}
  \caption{(color online) CPU utilization (top), GPU utilization (middle), runtime (bottom) as a function of \texttt{OMP\_NUM\_THREADS} for the TensorFlow (orange), PyTorch (blue) and NumPy (green) simulations as defined in Table~\ref{tab:simulation-setup}.}
  \label{figure:omp-results}
\end{figure}

The simulation using \texttt{NumPy} acts as reference simulation that does not involve any GPU specific libraries. For the simulations using \texttt{PyTorch} and \texttt{TensorFlow} we adjusted the number training steps in such a way that the simulations without any limitation have a runtime of roughly 30 minutes, logging various information during the simulations. In the following we focus on the number of threads (Fig.~\ref{figure:threads}), CPU usage, GPU usage and runtime (Fig.~\ref{figure:omp-results}).

The main purpose of environment variables like \texttt{OMP\_NUM\_THREADS} is to control the number of threads (NoT) a program uses, therefore this is the first variable to have a closer look at, Fig.~\ref{figure:threads}.

Without any restrictions, the NoT for \texttt{TensorFlow} is $\mathrm{NoT}=80$, for \texttt{PyTorch} $\mathrm{NoT}=9$ and for \texttt{NumPy} $\mathrm{NoT}=1$. For all combinations of the environment variables defined in Algorithm~\ref{lst:envtest}, we do not see any change in NoT for \texttt{NumPy} and \texttt{PyTorch}. Whereas for \texttt{TensorFlow} the NoT varies between $57$ and $72$. Regarding \texttt{NumPy} and \texttt{PyTorch}, both do not use libraries that are sensitive to any of the used environment variables, meaning the appearance of the threads is not related to a class or function defined in \texttt{libc}. In the case of \texttt{TensorFlow}, at least some of the threads can be controlled but this number is marginal compared to the NoT that cannot be controlled.

This result raises the question if the environment variables defined in Algorithm~\ref{lst:envtest} have any effect on the performance at all? As shown in Fig.~\ref{figure:omp-results}, there is no difference within errors in CPU usage or GPU usage and runtime.

There are situations in which we can use environment variables to control the number of threads even within \texttt{Python} scripts, as long as the appropriate libraries are used. For the most prominent DL frameworks -- \texttt{TensorFlow} and \texttt{PyTorch} -- this is not the case. This means one has to find other ways to control \texttt{Python} scripts or scripting languages in general. In the next section we have a closer look at one possibility which is \texttt{taskset}.

As it is the case for control groups we can use \texttt{taskset} to pin processes to specific cores. In contrast to control groups, every user can use \texttt{taskset}. We used the same \texttt{PyTorch} and \texttt{TensorFlow} setup as defined in Table~\ref{tab:simulation-setup} and omit the \texttt{NumPy} simulations. In detail we pinned the simulations to $1$, $4$, $8$ and $20$ CPU cores without any further limitations and in addition did the same again with setting \texttt{OMP\_NUM\_THREADS=1}. We do this extra restriction in order to detect or exclude the possibility of interactions between the pinning via \texttt{taskset} and environment variables.

\begin{figure}[t]
  \centering
  \includegraphics[width=\columnwidth]{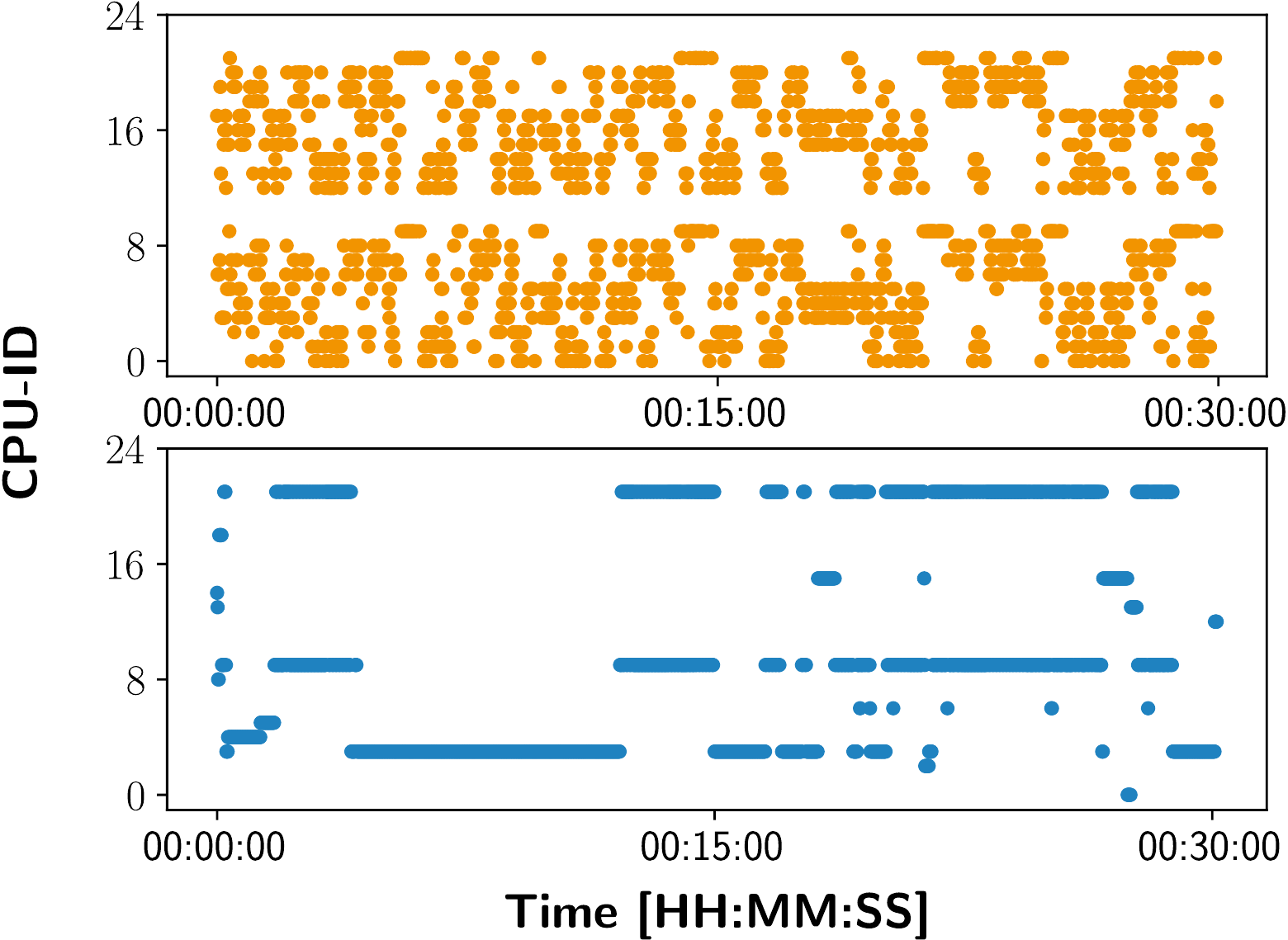}
  \vspace*{-0.25cm}
  \caption{(color online) CPU ID of the main process as a function of runtime of the TensorFlow (orange) and PyTorch (blue) simulations as defined in Table~\ref{tab:simulation-setup} without any restrictions. (Note that the group like splitting is because of the fact that $4$ CPU cores are blocked for the operating system of the compute node.)}
  \label{figure:taskset-main-ID}
\end{figure}

Before we go into details regarding the explicit pinning of the processes, we have a look at the CPU that the main process of the simulations uses as a function of time without any restrictions, Fig.~\ref{figure:taskset-main-ID}.

First of all, it is interesting to see that for both \texttt{TensorFlow} and \texttt{PyTorch} the used CPU cores of the main process changes over time. In the case of \texttt{TensorFlow}, the used CPU core stays sometimes for less that $5$ seconds the same. For \texttt{PyTorch} in contrast there are longer periods with the same CPU core. If we use \texttt{taskset} ,and pin the simulation to specific cores, the main process can only switch been cores it is pinned to.

\begin{figure}[t]
  \centering
  \includegraphics[width=\columnwidth]{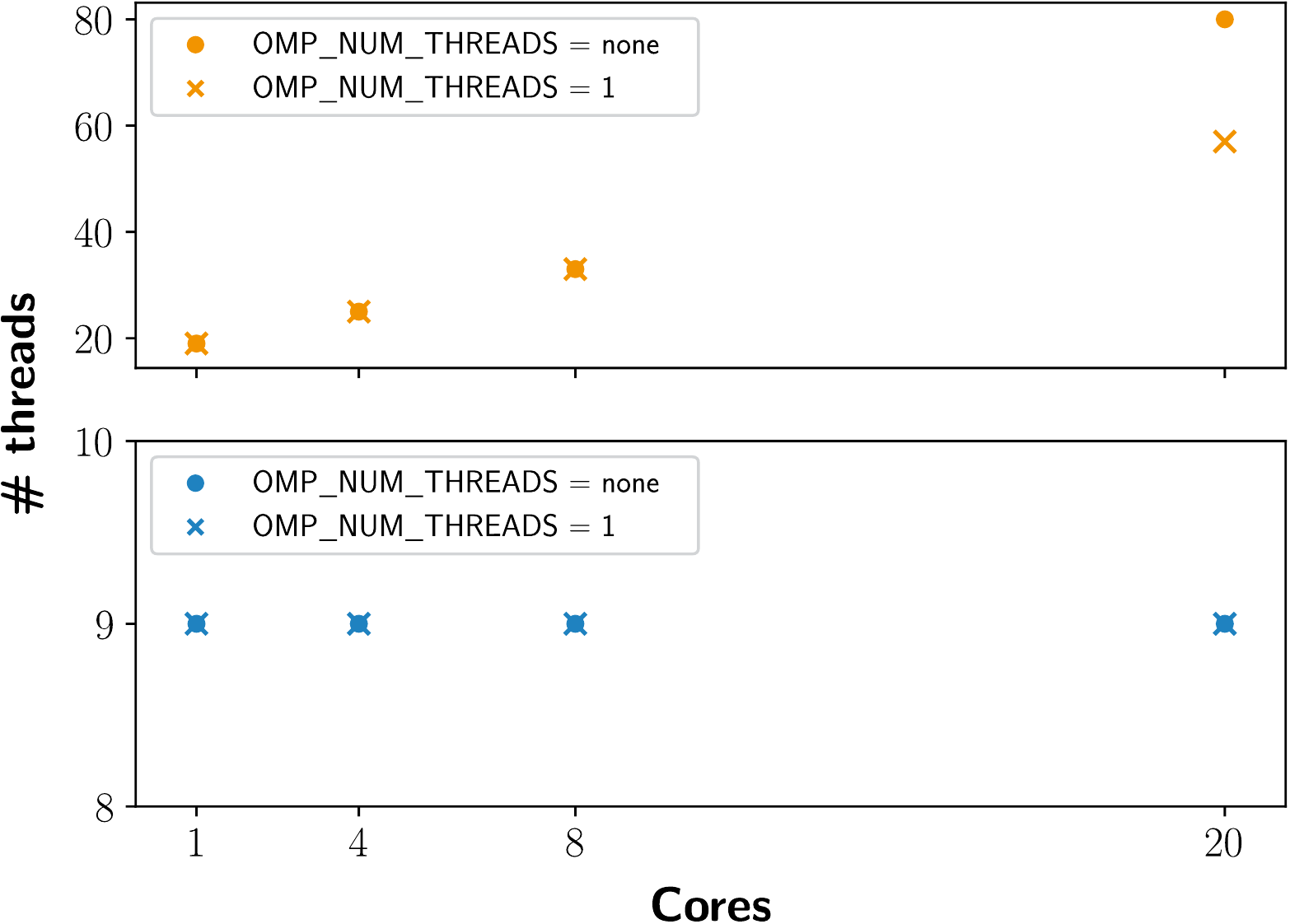}
  \vspace*{-0.25cm}
  \caption{(color online) NoT as a function the available CPU cores of the \texttt{TensorFlow} (orange) and \texttt{PyTorch} (blue) simulations as defined in Table~\ref{tab:simulation-setup}.}
  \label{figure:taskset-not}
\end{figure}

A similar behavior can be found for the NoT, Fig.~\ref{figure:taskset-not}. Regarding the details we have to be a bit restrictive.

In the case of \texttt{PyTorch} (blue symbols in Fig.~\ref{figure:taskset-not}) the NoT is not effected by the number of available cores as it was in Fig.~\ref{figure:threads}. For \texttt{TensorFlow} we see that the NoT is effected by the number of available cores. Nevertheless the lowest NoT -- during the training -- that can be reached is approximately $19$. In addition we see no difference between no restrictions and \texttt{OMP\_NUM\_THREADS=1} for $1$, $4$ and $8$ cores. Whereas for $20$ cores there is a difference in the two settings. In order to understand it is necessary to have a detailed look at the source code implementation of \texttt{TensorFlow}. Note that this is not the perspective of this paper and we would leave this open for future ongoing research.

\begin{figure}[t]
  \centering
  \includegraphics[width=\columnwidth]{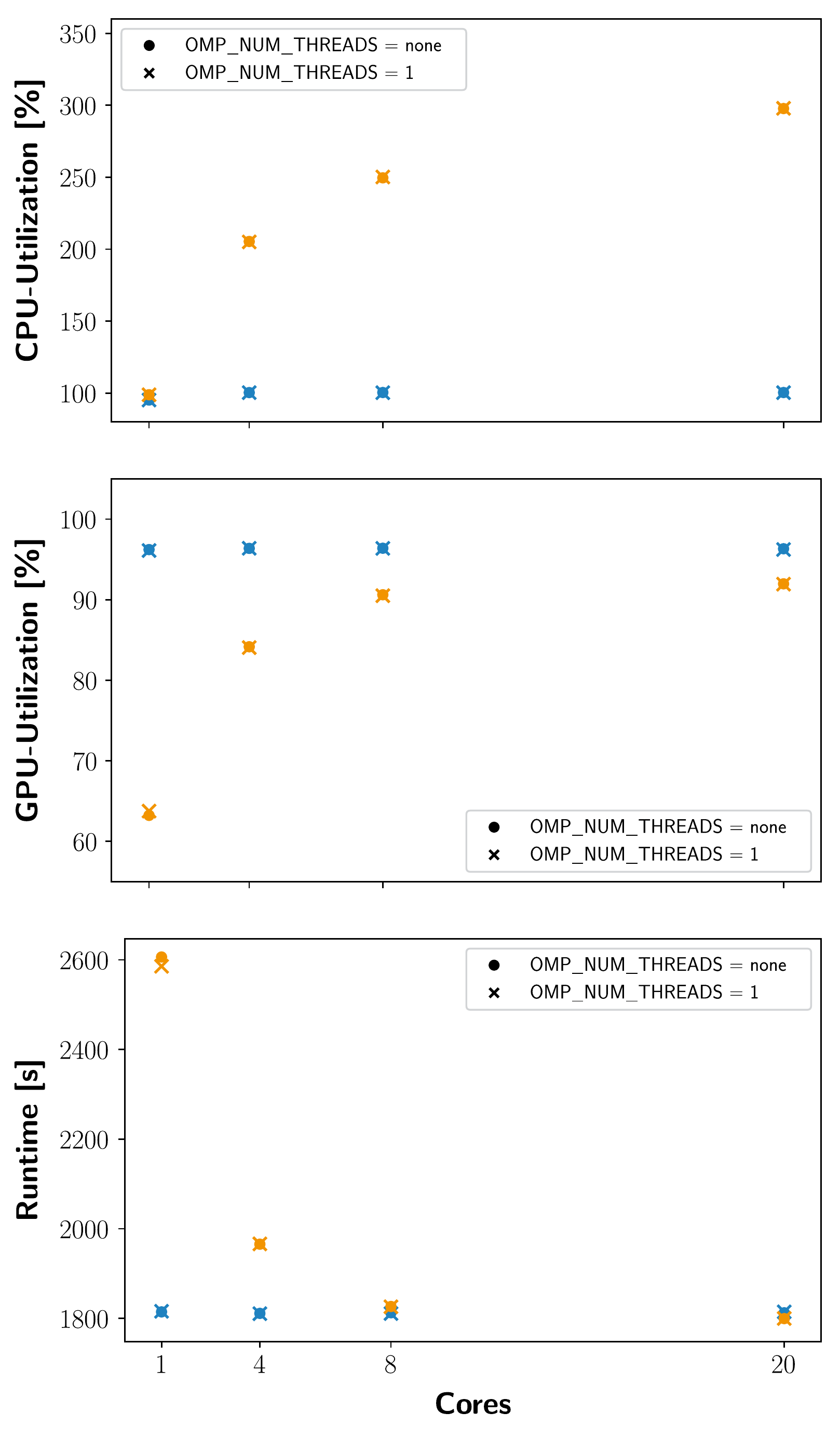}
  \vspace*{-0.25cm}
  \caption{(color online) \textit{CPU usage} (top), \textit{GPU usage} (middle) and \textit{runtime} (bottom) as a function of the available CPU cores of the \texttt{TensorFlow} (orange) and \texttt{PyTorch} (blue) simulations as defined in Table~\ref{tab:simulation-setup}.}
  \label{figure:taskset-results}
\end{figure}

From the view of a user it is perhaps even more interesting to know if limiting the number of cores has a significant effect on the performance or not. Even though most DL users focus on the GPU usage -- Fig.~\ref{figure:taskset-results}~(middle) -- the CPU usage is equally important, Fig.~\ref{figure:taskset-results}~(top).

For \texttt{PyTorch} it makes no difference if we have $1$ or $20$ cores available the CPU usage stays at $100\%$. Note that we talk about a straight forward implementation with default settings and without explicit parallelization. Like before, the \texttt{TensorFlow} implementation is more sensitive. For one available CPU core, the CPU usage stays at $100\%$ which proves that \texttt{taskset} works as supposed. With increasing number of cores the CPU usages increases as well but seems to saturate at roughly $300\%$. With GPUs present the CPU does not focus on heavy calculations but has the main focus on preprocessing and similar tasks. This raises the question if the GPU is effected by the CPU usage and in deed it is, Fig.~\ref{figure:taskset-results}~(middle). To be precise, it is the case for \texttt{TensorFlow}, for \texttt{PyTorch} we see no effect at all. Regarding \texttt{TensorFlow}, we find a nearly linear correlation between GPU and CPU usage. If we have only one CPU core the GPU usage drops to $63\%$ in average. With increasing number of cores the usage grows and reaches a saturation at approximately $90\%$.

As both CPU and GPU usage show partially strong differences depending on the number of available CPU cores we find that the runtime is affected as well, see Fig.~\ref{figure:taskset-results}~(bottom). Regarding the interpretation of the runtime, we have to be a bit more careful. In the case of \texttt{PyTorch} there is no effect as we see no effect in CPU or GPU usage as well. The runtime for the \texttt{TensorFlow} simulations resembles the results of the CPU and GPU usage. As the entire preprocessing is done on the CPU the less cores are available the harder it is too feed the GPU in a proper manor. This results in longer runtimes for less cores. Besides this it is not the case that the simulation time would continue to decrease with even more cores.

\begin{figure}[t]
  \centering
  \includegraphics[width=\columnwidth]{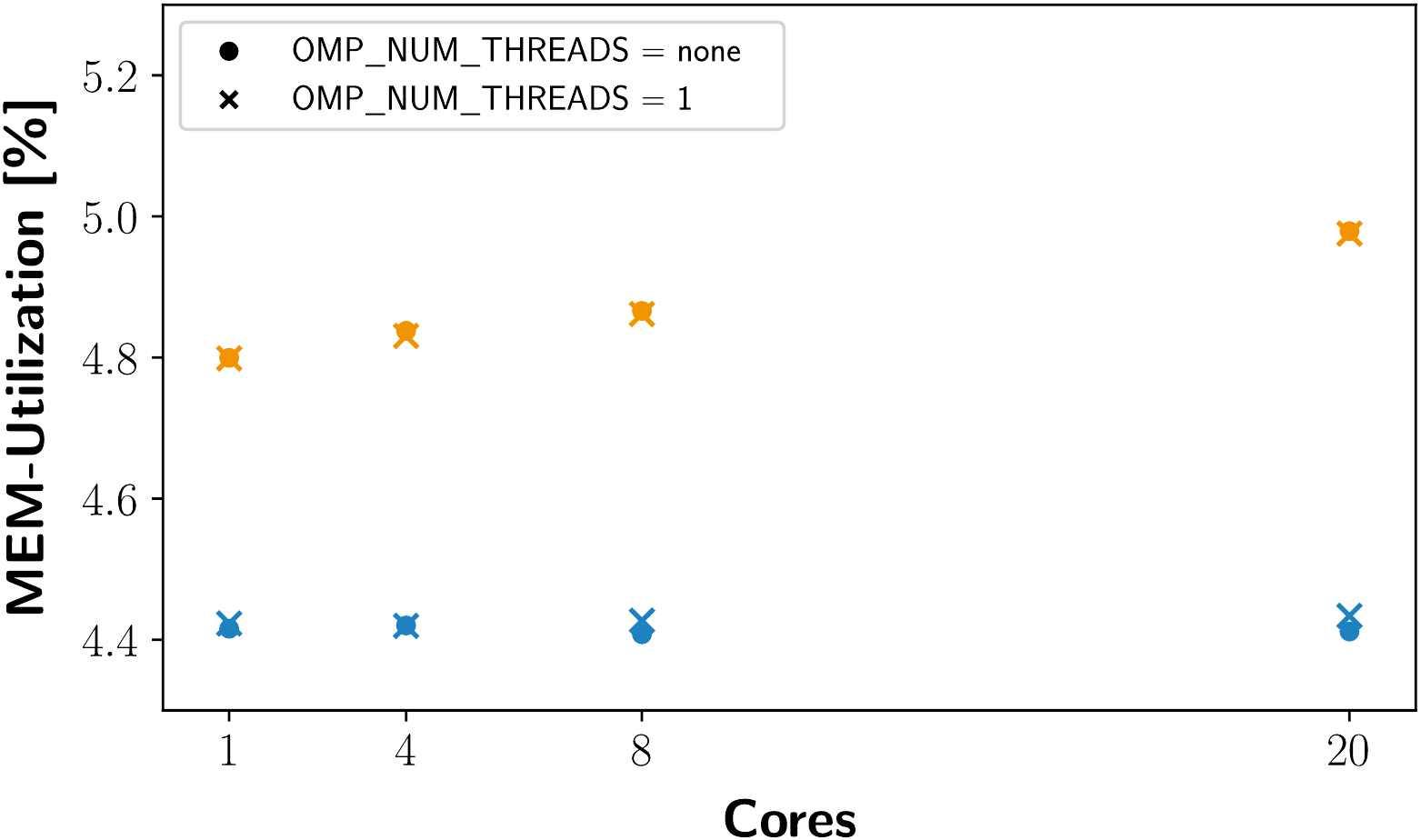}
  \vspace*{-0.25cm}
  \caption{(color online) \textit{Main Memory} as a function of the available CPU cores of the \texttt{TensorFlow} (orange) and \texttt{PyTorch} (blue) simulations as defined in Table~\ref{tab:simulation-setup}.}
  \label{figure:taskset-results-mem}
\end{figure}

Although we find effects in the case of CPU or GPU usage, the used main memory is, for both \texttt{PyTorch} and \texttt{TensorFlow}, nearly not effected at all, see Fig.~\ref{figure:taskset-results-mem}.

\section{Management of Python Environments}\label{sec:python-environment}
In order to provide a useful and manageable \texttt{Python} environment on HPC systems one has to rethink the common way software and libraries are installed. Essentially there are two main problems -- extremely fast changing software releases and dependency issues. The latter problem has rather diverse reasons. First of all we want to make the clear statement that \texttt{Python~2} and \texttt{Python~3} are two independent languages. Because of this it is not possible to execute scripts that contain \texttt{Python~2} code with a \texttt{Python~3} interpreter and vice versa. For ML, DL and DS it is standard that different methods or models have totally different dependencies which makes it impossible installing them at the same time on all compute nodes of a cluster. As a consequence we have to find another approach.

\subsection{Installing Python Software Stacks}
As most of the ML/DL community uses \texttt{Python} as main programming language, the usage of \textit{Anaconda Distribution}~\cite{anaconda} and/or \texttt{pip}~\cite{pip} to install needed packages and libraries is widespread. These make it possible for users to install essentially everything in their respective home directories. On the one side, this may be an advantage for the users as they can easily install things they need without explicitly asking for example system administrators. On the other side, this has two major drawbacks. First, it makes it hard to control what users install, which can end up with serious security issues. Depending on the provider of the respective cluster -- university, research institute or industry -- the legal consequences can be tremendous. Second, a local Anaconda installation for every user has about $5$GB mostly redundant data. We say ``redundant'' as there is a huge amount of libraries and binary files that many users need or is installed by anaconda on default. Despite the drawbacks there are advantages as well, e.g. multiple python environments (\texttt{Python~2} and \texttt{Python~3} simultaneously) and faster updates than one has in most Linux repositories (which is most of the time needed by ML/DL users).

\subsection{Maintaining Python Software Stacks}
In order to make it possible for users to get the advantages and have a better control over what is installed, we can make use of different approaches. For the redundancy part there is the possibility to install Anaconda in such a way that one gets multi-user support. In case of security concerns, one can create a \textit{local} Anaconda repository -- or in ``Anaconda language'' creating custom channel. With such a local repository system administrators are able to control the packages that are allowed on the system. Note that this follows the idea of a local clone of a Linux repository -- a well-known ``trick'' on HPC systems.

Besides the already mentioned difficulties with \texttt{Python} libraries, it is likely that there are more restrictions, e.g. different program versions for different users, more diverse security issues or regimentation in connection with the General Data Protection Regulation~\cite{GDPR}. One way to realize such requirements is the usage of virtualization tools, like \textit{virtual machines} or \textit{software containers}\footnote{Note that the goal of this paper is not to distinguish between container software like \textit{Singularity}~\cite{singularity} or \textit{Docker}~\cite{docker} or giving an overview of available container solutions. Regarding this we would like to recommend Ref.~\cite{cloudnative}.}. Depending on the particular use case and the cluster environment settings there are solutions that fit better. For example if we already have an existing batch system, e.g. Slurm~\cite{slurm} or LSF~\cite{lfs}, the preferred container solution has to make it possible to start containers as normal users.

\subsection{Challenges with GPU Drivers}
The perhaps most important aspect regarding DL is the native and easy GPU implementation and usage. The tricky part with the implementation is twofold. The GPU has to be available inside the container. This visibility goes hand in hand with the respective driver of the installed GPU. At the moment we have the situation for container solutions that the GPU driver has to be installed on the host system as it can otherwise not be available inside the container. This meant for a long time that inside the container was only one specific \texttt{CUDA} and therefore \texttt{TensorFlow} or \texttt{PyTorch} version possible. As a consequence one would have to boot the host system with another GPU driver to have another \texttt{CUDA} version inside the Container. This is not feasible on a productive cluster, as it would have a massive impact on both the workload and other users.
Starting with \texttt{CUDA} version $10$, it is \textit{now} possible to update both \texttt{CUDA} \textit{and} the GPU driver without touching the respective kernel modules.~\cite{nvidia-cuda} Therefore the remaining ``limiting part'' is the GPU driver. But this simplifies the situation a lot as we can now simply bind mount the libraries provided by the GPU driver in the container and install the different \texttt{CUDA}, \texttt{TensorFlow} and \texttt{PyTorch} versions directly in the container. This gives the user full control over which framework \textit{and} version he wants to use during his training.
Depending on the cluster, the used Linux distribution and the update policies on the cluster it might take some time until this feature will be available to all users. But the needed tools and software are available and it opens the door to the HPC world for the entire machine and deep learning community without making HPC clusters ``useless'' for other researchers.

\section{Conclusion}\label{sec:conclusion}
The rise of compute intensive machine learning and data analytics applications has a big impact on the design and maintenance of HPC-Systems: the increased usage of GPUs on the hardware side, and the application of \texttt{Python} workflows on the software side. In combination with new user groups, which have little HPC background, these factors can have a massive impact on the operation of a HPC-cluster. The problems that we have discussed throughout this paper mostly originate directly or indirectly from the use of \texttt{Python} environments. As we have shown, most of these problems are solvable, but need some attention until HPC batch-systems and other parts of the HPC software stack have evolved to handle them automatically.

The most promising solution towards a user friendly, secure and efficient adaption of \texttt{Python} workflows on HPC systems appears the usage of containers. This aspect has not been part of the presented investigations, but we are planing to extend our work in this direction.



\end{document}